%% file: main.tex
\pdfoutput=1

\documentclass[11pt]{article}

\usepackage[]{acl}

\usepackage{times}
\usepackage{latexsym}

\usepackage[T1]{fontenc}

\usepackage[utf8]{inputenc}

\usepackage{microtype}
\usepackage{subfigure}
\usepackage{graphicx}
\usepackage{makecell}
\usepackage{multirow}
\usepackage{amsmath}
\usepackage{xspace} 
\usepackage{CJKutf8} 
\usepackage{hyperref}
\newcommand\data{\mbox{SSD}\xspace}
\newcommand\datap{\mbox{SSD-{PHONE}}\xspace}
\newcommand\datai{\mbox{SSD-{ID}}\xspace}
\newcommand\datan{\mbox{SSD-{NAME}}\xspace}
\newcommand\datac{\mbox{SSD-{PLATE}}\xspace}
\newcommand\task{\mbox{SSTOD}\xspace}
\newcommand\our{\mbox{UBAR$^{+}$}\xspace}
\newcommand{\tabincell}[2]{\begin{tabular}{@{}#1@{}}#2\end{tabular}} 
\newcommand\zh[1]{\begin{CJK*}{UTF8}{gbsn}#1\end{CJK*}}

\title{A Slot Is Not Built in One Utterance: Spoken Language Dialogs with Sub-Slots}

\author{
    Sai Zhang \textsuperscript{\rm 1}{\thanks{\ \ Equal contribution. }},
    Yuwei Hu\textsuperscript{\rm 1}{\footnotemark[1]},
    Yuchuan Wu\textsuperscript{\rm 2},
    Jiaman Wu\textsuperscript{\rm 2}, \\
    {\bf Yongbin Li\textsuperscript{\rm 2}\thanks{ \ \ Yongbin Li is the corresponding author.}},
    {\bf Jian Sun}\textsuperscript{\rm 1},
    {\bf Caixia Yuan\textsuperscript{\rm 1}}
    {\bf \and Xiaojie Wang\textsuperscript{\rm 1}} \\
    \textsuperscript{\rm 1}Beijing University of Posts and Telecommunications, Beijing, China\\
    \textsuperscript{\rm 2}Independent Researcher \\
    \texttt{\{zs, hyw724, yuancx, xjwang\}@bupt.edu.cn} \\
    \texttt{jw.0123@outlook.com} , \texttt{jiansun\_china@hotmail.com} \\
    \texttt{liyb821@gmail.com}
}

\begin{document}
\maketitle
\begin{abstract}
    \input{0_abs}
\end{abstract}
\input{1_intro}
\input{2_Task_and_Data}

\input{3_Method}
\input{4_Experiments}
\input{6_Related_Work}
\input{7_Conclusion}

\bibliographystyle{acl_natbib}
\bibliography{ref}

\clearpage
\appendix
\input{8_Appendix}

\end{document}

%% file: 0_abs.tex
A slot value might be provided segment by segment over multiple-turn interactions in a dialog, especially for some important information such as phone numbers and names. It is a common phenomenon in daily life, but little attention has been paid to it in previous work. To fill the gap, this paper defines a new task named \textbf{S}ub-\textbf{S}lot based \textbf{T}ask-\textbf{O}riented \textbf{D}ialog (\textbf{\task}) and builds a Chinese dialog dataset \data for boosting research on \task. The dataset includes a total of 40K dialogs and 500K utterances from four different domains: Chinese names, phone numbers, ID numbers and license plate numbers. The data is well annotated with sub-slot values, slot values, dialog states and actions. We find some new linguistic phenomena and interactive manners in \task which raise critical challenges of building dialog agents for the task. We test three state-of-the-art dialog models on \task and find they cannot handle the task well on any of the four domains. We also investigate an improved model by involving slot knowledge in a plug-in manner. More work should be done to meet the new challenges raised from \task which widely exists in real-life applications. The dataset and code are publicly available via \href{https://github.com/shunjiu/SSTOD}{https://github.com/shunjiu/SSTOD}.

%% file: 1_intro.tex
\section{Introduction}

\begin{figure}
    \centering
    \includegraphics[width=\linewidth]{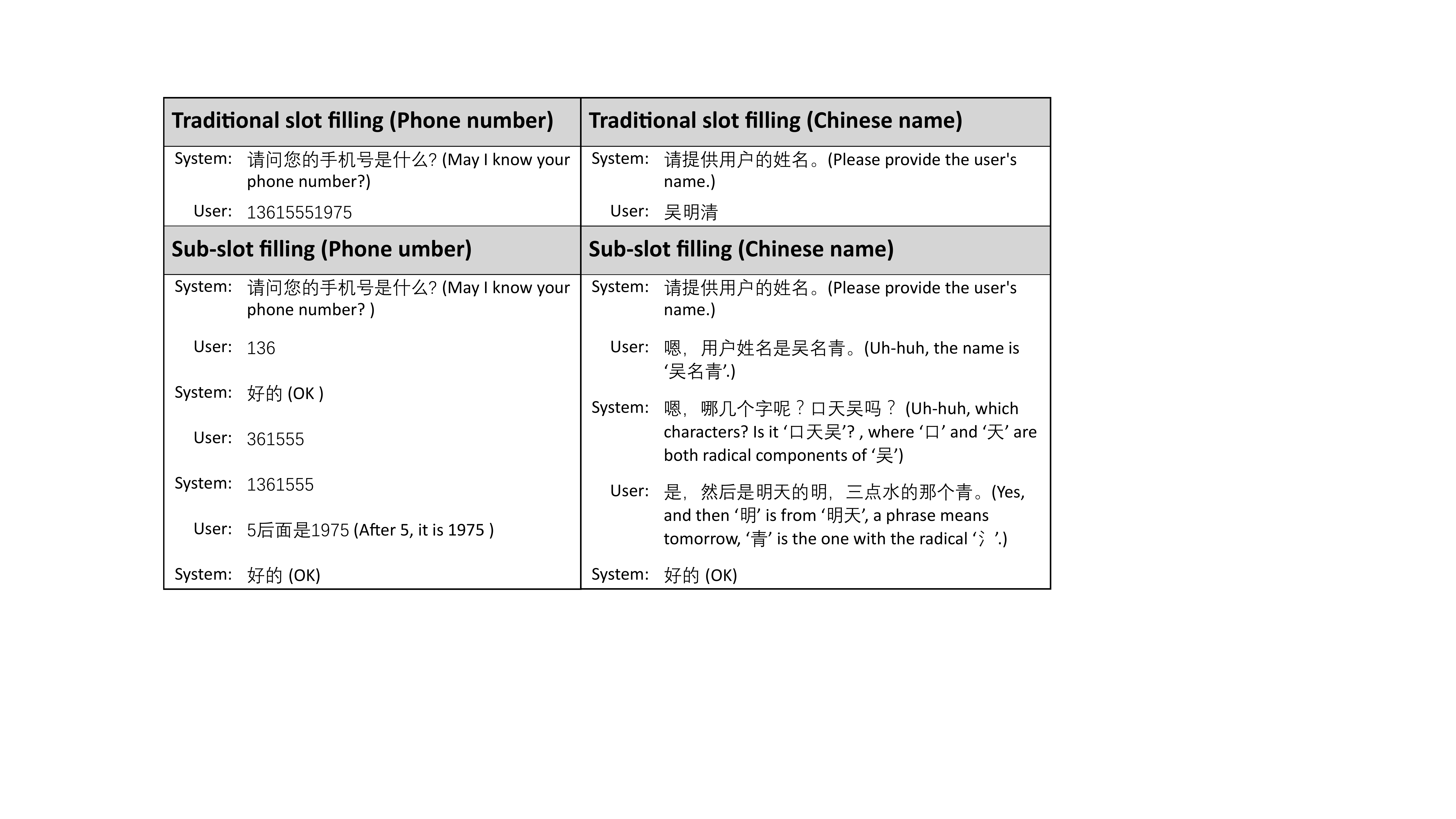}
    \caption{Comparison of traditional slot and sub-slot.}
    \label{fig: example of SSD}
\end{figure}

Task-oriented dialogs help users accomplish specific tasks such as booking restaurants or accessing technical support services by acquiring task-related slots through multi-turn dialogs. Many advances have been achieved under an assumption that each slot value is informed or updated as a whole in a single turn by default ~\citep{Li-2017-tod,z.zhang-2020,2020-simpleTOD, dai-etal-2021-preview}. But in real-world dialogs, some slot values are often provided in a much more complicated manner. We take phone numbers as an example. Users tend to inform an agent a sequence of 0-9 digits segment by segment across several turns as exemplified in Figure~\ref{fig: example of SSD}. Accordingly, the agent needs to confirm, update or record the recognized sub-slot values. We regard these scenarios as \task task.

The \task is very common when people communicate telephone numbers, names and so on. Specifically, as shown in Figure~\ref{fig: example of SSD}, the \task task raises several critical new challenges which have not been tackled in building dialog agents: (1) Multi-segment informing: The segments could be informed in many different complex ways. As exampled in Figure ~\ref{fig: example of SSD}, the user informed two sub-slots ``136'' and ``361555'' sequentially. The agent should discriminate whether the snippet ``36'' in ``361555'' is a partial repeat of  ``136'' or a duplicated component of a whole slot value. (2) Sub-slot locating: Differing from updating a whole slot value in traditional slot filling, in \task, the agent demands to precisely locate the part of values that needs to be updated. The situation is exacerbated when there are more than one similar sub-slots. (3) Knowledge-rich relevancy: To avoid the ambiguities of speech, users usually introduce a piece of knowledge along with informing the slot values~\citep{CDR,wang2007interactive}. For example, the knowledge, ``\begin{CJK*}{UTF8}{gbsn}明天的明\end{CJK*}'' is used to disambiguate character ``\begin{CJK*}{UTF8}{gbsn}明\end{CJK*}'' (It is the similar case when English speakers say ``A as in Alpha'' in phone calls). The agent should look into the knowledge in order to predict correct value. 

To the best of our knowledge, the existing dialog benchmarks, such as ATIS~\citep{hemphill-etal-1990-atis}, MultiWOZ~\citep{budzianowski-etal-2018-multiwoz}, CrossWOZ~\citep{zhu-etal-2020-crosswoz}, and SGD~\citep{rastogi2020towards} do not contain the dialogs illustrated in Figure~\ref{fig: example of SSD}, which makes the dialog agents optimized on them fail dramatically at conversing in sub-slot dialogs. To address the above challenges, we develop the \textbf{S}ub-\textbf{s}lot \textbf{D}ialog (\data) dataset which contains most popular sub-slot dialog scenarios including phone numbers (a sequence of digits 0-9), ID numbers (much longer digit sequence), person names (a sequence of Chinese characters), and license plate numbers (a mix of Chinese characters, digits and English letters). The dataset is originated from the real-world human-to-human conversations, then richly labeled and reprocessed by crowdsourcing. Although the dataset is in Chinese, the development methodology depicted in this work is also applicable to other languages. 


Under the setting of \task, we present an improved model, \our, on the basis of UBAR~\citep{yang2021ubar} and the large pretrained model GPT2~\citep{radford-2019-gpt2}. \our equips UBAR with a knowledge prediction module to correct Automatic Speech Recognition (ASR) errors and discriminate the ambiguities, and achieves better  performance on \data. We also provide a rule-based user simulator to evaluate the system.

Our main contributions are:
\begin{itemize}
    \item We propose a novel sub-slot based dialog task which exists widely in real-world conversations but has been neglected in previous work.
    \item We build a large-scale high-quality spoken Chinese dataset \data for \task, covering four common scenarios including phone numbers, ID numbers, Chinese names and license plate numbers collection, which will essentially benefit future research on \task.
    \item We design a knowledge prediction module together with knowledge retrieval which helps UBAR achieve significant improvement on the name domain. Otherwise, a user simulator is provided to facilitate the evaluation of the system.
\end{itemize}

%% file: 2_Task_and_Data.tex
\section{Task and Dataset}

We first give a defination of SSTOD, then introduce how to build the \data dataset, and give some analyses on the dataset. 

\subsection{Task Defination}

We proposed sub-slot based dialog system as a one slot filling task. A user may provide a slot via multiple turns in oral conversations. In each turn, only a piece of the value, which is regarded as a sub-slot, is given. It is because the values like phone numbers are usually too long for a user to inform in one turn or the segments in values like surnames in names are often accompanied with extra explanations to disambiguate homonyms.

\subsection{Dataset Creation}

Since information such as phone numbers and names is private, real data cannot be used directly. We design a semi-automatic method to obtain a large-scale high-quality dialog dataset while avoiding privacy issues. We build a dataset in four domains including Phone Number, Name, ID Number and License Plate Number. We demonstrate the building process of the dataset by taking Name as an example. 

\begin{figure}[ht]
    \centering
    \includegraphics[width=\linewidth]{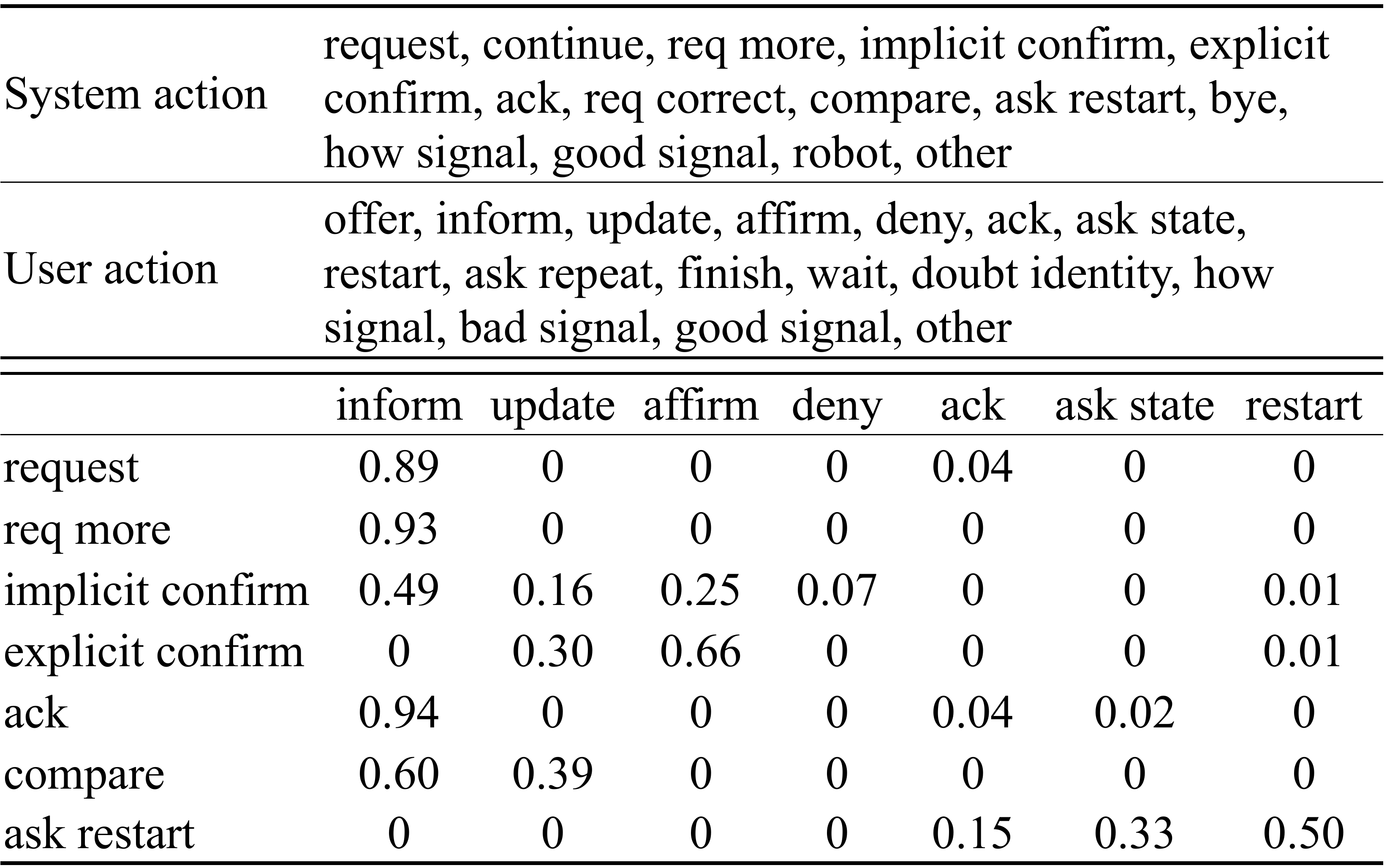}
    \caption{All actions in the phone domain (above) and part of transition probabilities (below). Each row in the table below is the probability of user action when a system action is given.}
    \label{fig: act jump}
\end{figure}

\begin{figure*}[ht]
	\centering
	\begin{minipage}[t]{0.43\linewidth}
    	\centering
    	\includegraphics[width=\linewidth]{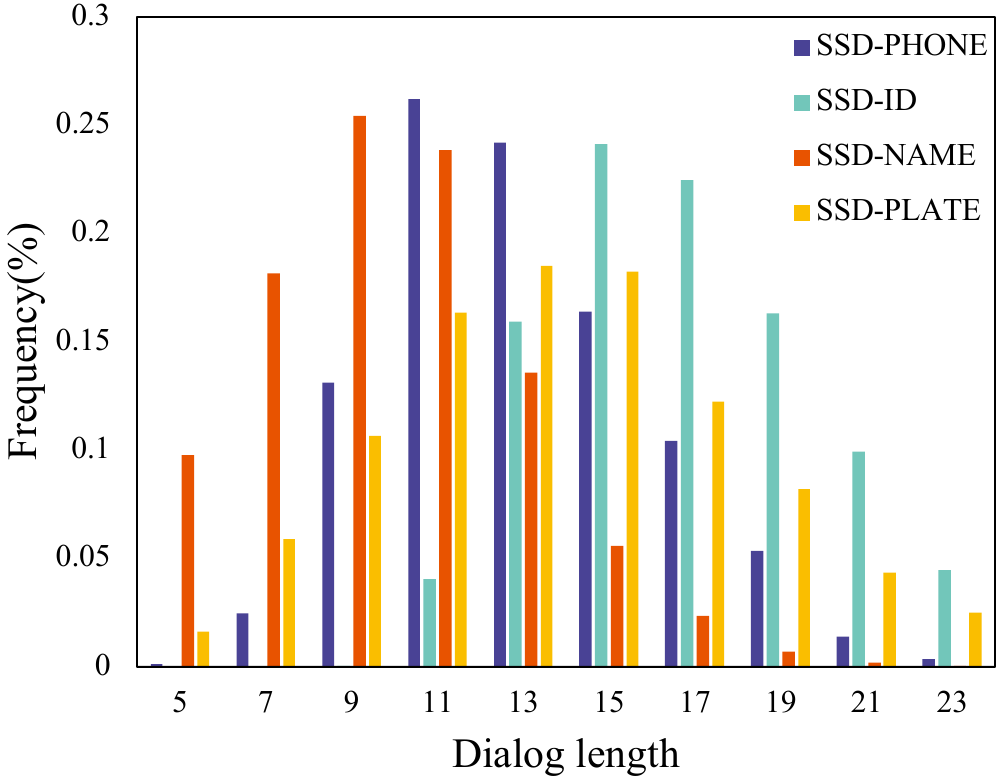}
    \end{minipage}
	\begin{minipage}[t]{0.49\linewidth}
    	\centering
    	\includegraphics[width=\linewidth]{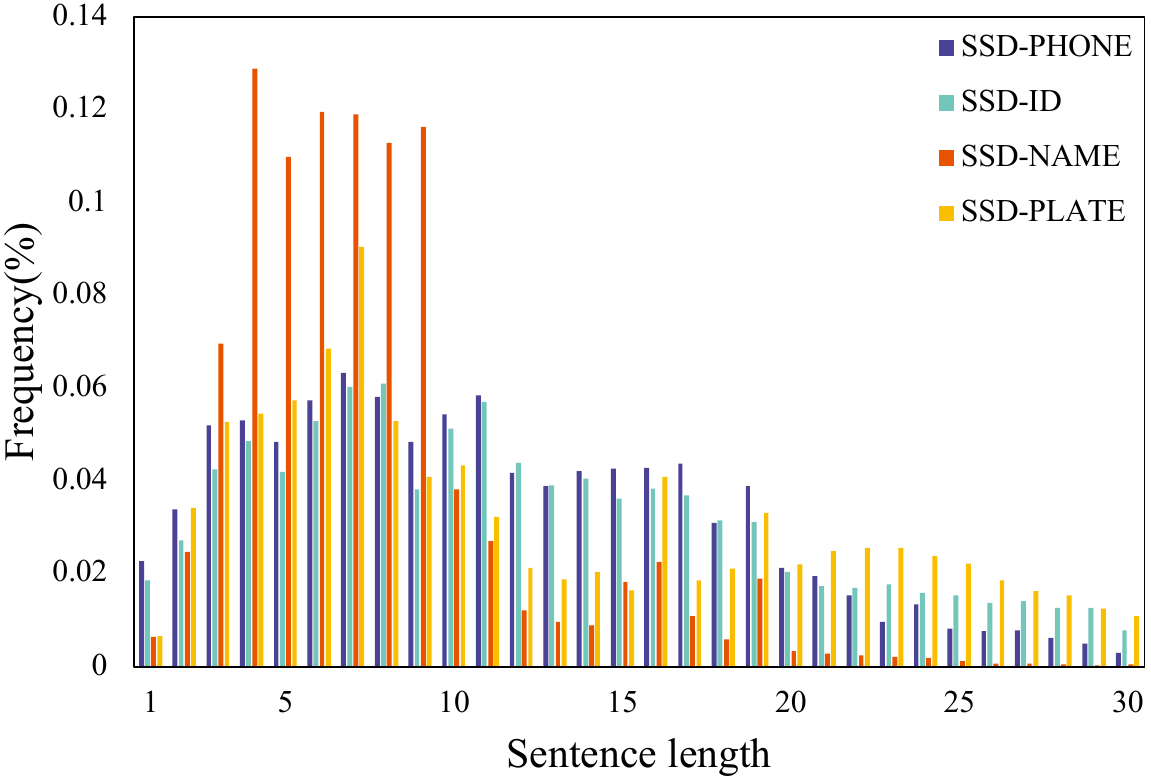}
    \end{minipage}
    \caption{The distribution of numbers of sentences in a dialog (left) and the distribution of numbers of characters in a sentence (right).}
    \label{fig:dialogucount}
\end{figure*}

\paragraph{Human-to-Human (H2H) dialog.} We sample $47,252$ H2H dialogs from a business service by considering different time of service and different genders of customers, and obtain $4,489$, $8,873$ and $5,827$ fragments of dialog for phone numbers, names and license plate numbers respectively. We analyze the H2H dialogs carefully, summarize some dialog actions and dialog policy, and estimate the transition probabilities between different actions. Taking phone numbers as an example, we have $30$ actions. Figure~\ref{fig: act jump} gives part of transition probabilities between those actions.

\paragraph{Knowledge Base.} Chinese characters in names cannot be disambiguated by context in spoken conversations. For example, when someone says, ``\begin{CJK*}{UTF8}{gbsn}我姓吴\end{CJK*} (my surname is Wu)'', different Chinese characters which share the same pronunciation of ``wu'', including ``\begin{CJK*}{UTF8}{gbsn}吴\end{CJK*}'', ``\begin{CJK*}{UTF8}{gbsn}武\end{CJK*}'', ``\begin{CJK*}{UTF8}{gbsn}伍\end{CJK*}'', etc., are all possible to be the surname to the listeners. People therefore always employ some external knowledge to distinguish different characters. For example,  ``\begin{CJK*}{UTF8}{gbsn}我姓吴,口天吴\end{CJK*} (my surname is `\zh{吴}', `\zh{口}' and `\zh{天}' compose `\zh{吴}')'', where ``\zh{口天吴}'' is a piece of external knowledge. It gives components (normally some simple characters) of a character. People also use knowledge of character combination (i.e. words or phrases) to identify a Chinese character. For example, ``\begin{CJK*}{UTF8}{gbsn}我姓吴,东吴的吴\end{CJK*} (my surname is `Wu', `Wu' as in `DongWu') '', where ``DongWu'' is a word which only  ``\begin{CJK*}{UTF8}{gbsn}吴\end{CJK*}'' fits the word well. ``DongWu'' is another piece of knowledge for Chinese character ``\begin{CJK*}{UTF8}{gbsn}吴\end{CJK*}''.
Almost all frequent Chinese Characters have several pieces of knowledge as above. Appendix~\ref{sec:appendix1} gives some pieces of knowledge on Chinese characters. 
Knowledge is widely used in name telling. We thus build $20,547$ pieces of knowledge for $2,003$ common used Chinese characters. On average, each Chinese character is with more than $10$ pieces of knowledge. We give more examples in Appendix~\ref{sec:appendix1}.

\paragraph{Data generation.} Based on the analysis of H2H dialogs, two probabilistic FSA-based simulators are built for System and User respectively, both with a template-based Nature Language Generation (NLG) module for generating natural language sentences from actions sampled from probabilistic FSA. We give some examples of NLG modules in Appendix~\ref{sec:appendix2}. Part of FSAs is given in Appendix~\ref{sec:appendix3}. An error simulator is also built for modeling errors brought by ASR. 
Two FSAs as well as a NLG module and an error model work together to generate various dialogs. At the beginning, the FSA for users initializes a target slot value which is composed of several sub-slot segments. The two probabilistic FSAs then interact based on the sampled actions. At each step, when FSA chooses current dialog action and sub-slot values, a NLG template is randomly chosen to generate a sentence. The error model might also be triggered randomly to twist the values with a defined probability. When the system thinks it collects a complete slot value, it ends the dialog. If the slot value collected is consistent with the slot value initialized by the user, the dialog succeeds; otherwise, the dialog fails. Appendix~\ref{sec:appendix4} illustrates several example dialogs generated by FSAs.

\begin{table}[t]
    \centering
    \small
    \begin{tabular}{c|cccc} \hline
    \makecell[c]{Domains$\rightarrow$ \\ Types$\downarrow$ } & PHONE & ID & NAME & PLATE \\ \hline
    Templates & $8,578$ & $7,350$ & $3,031$ & $5,179$ \\
    Sentences & $3,849$ & - & $29,874$ & $10,000$ \\
    Knowledge & - & - & $34,302$ & - \\ \hline
    \end{tabular}
    \caption{Numbers of crowdsourced data.}
    \label{crowdsourced}
\end{table}

\begin{table*}[ht]
    \centering
    \begin{tabular}{c|cccc} \hline
     & \datap & \datai & \datan & \datac \\ \hline
    No. of dialogs & $11,000$ & $8,000$ & $15,000$ & $6,000$ \\
    No. of actions & $30$ & $30$ & $29$ & $27$ \\
    Avg. turns per dialog & $13.01$ & $16.86$ & $9.86$ & $13.90$ \\
    Avg. tokens per sentence & $11.61$ & $13.13$ & $7.70$ & $13.84$ \\
    Avg. sub-slots per dialog & $2.90$ & $4.15$ & $2.84$ & $2.03$ \\
    No. of different paths & $3,135$ & $5,412$ & $2,475$ & $3,965$ \\ 
    Vocabulary size & $677$ & $629$ & $3,519$ & $915$ \\\hline
    \end{tabular}
    \caption{Analysis of the \data dataset.}
    \label{ssd analyse}
\end{table*}

\paragraph{Data crowdsourcing.} To make our dialog data more natural and diverse, we hired crowd workers to paraphrase user utterances in the generated dialogs. New utterances bring more templates, knowledge pieces and real ASR errors. 
Table~\ref{crowdsourced} gives the numbers of crowdsourced data.

\subsection{Data Statistics}

We finally obtained a large and high-quality data for \task in four domains. Some statistics are shown in Table~\ref{ssd analyse}.

As we can seen in Table~\ref{ssd analyse}, the \data dataset has 40K dialogs and the number of dialogs exceeds that of most available task-oriented datasets (the largest dialog dataset SGD~\citep{rastogi2020towards} commonly used today contains $16,142$ dialogs). The number of actions is at least $27$ in each domain, which is more than that in any single domain of the currently commonly used dataset MultiWOZ~\citep{budzianowski-etal-2018-multiwoz}. 

The average turn per dialog is no less than $10$, as well as the average character per sentence. The distribution of dialog length is shown in Figure~\ref{fig:dialogucount} (left) and the distribution of dialog sentence length per domain is shown in Figure~\ref{fig:dialogucount} (right).

A path is the action sequence in a dialog. Two dialogs with distinct paths means they have different ways to complete a task. The larger the number of different paths, the more diversity of action sequences. The \data dataset shows adequate diversity of dialogs.

The average number of sub-slots per dialog is the average number of pieces that a full slot value is segmented. It can be seen that names are averagely segmented into $2.84$ pieces. Considering a Chinese name normally includes 2-3 Chinese characters, people say their names character by character.    

Finally, it should be noticed that data contains a wealth of annotation information. For each user utterance, we annotate an action and the sub-slot values provided by the user. For each system utterance, we annotate an action and the state which is the sub-slot value collected by the system. The annotation information allows our data to be used for the following tasks: natural language understanding (NLU), dialog state tracker (DST), dialog policy, NLG, etc. We will also release our FSA-based User simulator, which can be used to evaluate the system.

\begin{table*}[t]
    \small
    \centering
    \begin{tabular}{|l|p{1.15\columnwidth}|} 
    \hline
    \textbf{Description}  & \textbf{Example} \\ \hline
	Inform (quantifier) & \tabincell{l}{
		\begin{CJK*}{UTF8}{gbsn} 1，4个3 \end{CJK*}(1, four 3's.)
	} \\ \hline
	Inform (correct) & \tabincell{l}{
		\begin{CJK*}{UTF8}{gbsn} 嗯1820，呃，不是是1860 \end{CJK*}(Uh-huh1820, hmm, no it's 1860.)\\
	} \\ \hline
	Inform (repeat) & \tabincell{l}{
		\begin{CJK*}{UTF8}{gbsn} 7127 7127 \end{CJK*}
	} \\ \hline
	Inform (stretched) & \tabincell{l}{
		\begin{CJK*}{UTF8}{gbsn} 1，1044 \end{CJK*}
	} \\ \hline
    Inform (overlap) & \tabincell{l}{
		User: \begin{CJK*}{UTF8}{gbsn} 嗯，您那麻烦，您记一下的手机号码，181 \end{CJK*}(Well, would you \\ mind writing down the phone number? 181.)\\
		System: \begin{CJK*}{UTF8}{gbsn} 嗯，181 \end{CJK*}(Uh-huh, 181.)\\
		User: \begin{CJK*}{UTF8}{gbsn} 1814104 \end{CJK*}
	} \\ \hline
	Update (refer) & \tabincell{l}{
		\begin{CJK*}{UTF8}{gbsn} 最后4位是5664 \end{CJK*}(The last 4 digits are 5664.)
	} \\ \hline
	Update (delete) & \tabincell{l}{
		\begin{CJK*}{UTF8}{gbsn} 去掉7 \end{CJK*}(Delete 7.)
	} \\ \hline
	Update (add) & \tabincell{l}{
		\begin{CJK*}{UTF8}{gbsn} 9后面少个4 \end{CJK*}(Behind 9, 4 is missing.)
	} \\ \hline
    Update (part) & \tabincell{l}{
		System: \begin{CJK*}{UTF8}{gbsn} 133 4777 3029，好，我知道了，谢谢啊 \end{CJK*}(133 4777 3029, \\ okay, I see. Thanks!)\\
		User: \begin{CJK*}{UTF8}{gbsn} 529才对 \end{CJK*} (It is 529.)
	} \\ \hline
	Sub-slot update  & \tabincell{l}{
		\begin{CJK*}{UTF8}{gbsn} 2不对啊，是R，RST里面的R才对 \end{CJK*} (2 is not right, it's R as in RST. ) \\ (note: 2 and R have the same pronunciation in Chinese.) 
	} \\ \hline
	Comparison of homophonic characters & \tabincell{l}{
		\begin{CJK*}{UTF8}{gbsn} 是字母E还是数字1？ \end{CJK*}(	Is it the letter E or number 1? ) (note: ``E'' and \\ ``1'' have the same pronunciation in Chinese.) 
	} \\ \hline
    Using external knowledge (character combination)  & \tabincell{l}{
		\begin{CJK*}{UTF8}{gbsn} 艳是艳丽的艳 \end{CJK*} (``\begin{CJK*}{UTF8}{gbsn}艳\end{CJK*}'' is from ``\begin{CJK*}{UTF8}{gbsn}艳丽\end{CJK*}'', a two-character word means \\showy.)
	} \\ \hline
	Using external knowledge (structure)  & \tabincell{l}{
		 \begin{CJK*}{UTF8}{gbsn} 艳是一个丰字，一个色字 \end{CJK*} (``\begin{CJK*}{UTF8}{gbsn}艳\end{CJK*}'' is composed of ``\begin{CJK*}{UTF8}{gbsn}丰\end{CJK*}'' and ``\begin{CJK*}{UTF8}{gbsn}色\end{CJK*}''.)
	} \\ \hline
	ASR errors of a character or(and) its knowledge & \tabincell{l}{
		ASR outputs: \begin{CJK*}{UTF8}{gbsn} 验是严厉的严，一个风字，一个色字 \end{CJK*} \\
		Original utterance: \begin{CJK*}{UTF8}{gbsn} 艳是艳丽的艳，一个丰字，一个色字 \end{CJK*} \\
		(``\begin{CJK*}{UTF8}{gbsn}验\end{CJK*}'' and ``\begin{CJK*}{UTF8}{gbsn}严\end{CJK*}'' are badly recognized characters of ``\begin{CJK*}{UTF8}{gbsn}艳\end{CJK*}'',  ``\begin{CJK*}{UTF8}{gbsn}风\end{CJK*}'' is a \\badly recognized character of ``\begin{CJK*}{UTF8}{gbsn}丰\end{CJK*}'', and  ``\begin{CJK*}{UTF8}{gbsn}艳丽\end{CJK*}'' (showy) is the \\  correction of ``\begin{CJK*}{UTF8}{gbsn}严厉\end{CJK*}'' (servere).)
	} \\ \hline
	Two identical characters in one name  & \tabincell{l}{
		\begin{CJK*}{UTF8}{gbsn} 我叫李壮壮，状是状元的状，两个状都是 \end{CJK*}  (My name is ``\begin{CJK*}{UTF8}{gbsn}李壮壮\end{CJK*}'' \\ (Li Zhuangzhuang), the last two words are both ``\begin{CJK*}{UTF8}{gbsn}状\end{CJK*}'' as in ``\begin{CJK*}{UTF8}{gbsn}状元\end{CJK*}'' \\(top students).)
	} \\ \hline
	Two characters from one knowledge  & \tabincell{l}{
		\begin{CJK*}{UTF8}{gbsn} 我叫业勤，业精于勤的业勤 \end{CJK*} (My name is ``\begin{CJK*}{UTF8}{gbsn}业勤\end{CJK*}'' (Ye Qin) as in \\ Chinese idiom ``\begin{CJK*}{UTF8}{gbsn}业精于勤\end{CJK*}'' (Excellence in work lies in diligence).)
	} \\ \hline
    \end{tabular}
    
\caption{Part of the diversity cases and their examples.}
\label{tbl:diversity-example}
\end{table*}

\subsection{New Challenges}

The dataset includes many new phenomena that are seldom seen in other datasets which bring some new challenges to build agents for \task. Most of the new phenomena are brought by the sub-slot telling way. Table~\ref{tbl:diversity-example} gives some of these new phenomena as well as a sample utterance for each phenomenon.

Most of the phenomena listed in Table~\ref{tbl:diversity-example} are seldom seen in the previous dialog datasets. They raise some new challenges on at least three sides: The first one is to locate and record each segment and even each element in each segment, since all of them might be updated separately or as a whole. The second one is to identify the various external knowledge, especially when ASR errors are involved. The third one is that the context of the sub-slot might be helpless when there are ambiguities. The knowledge might be the major source of disambiguation, including those explicitly noticed in utterances, as well as implicitly used in dialogs.

%% file: 3_Method.tex
\section{Method}

\subsection{Benchmark Models}

\begin{figure*}
    \centering
    \includegraphics[width=\linewidth]{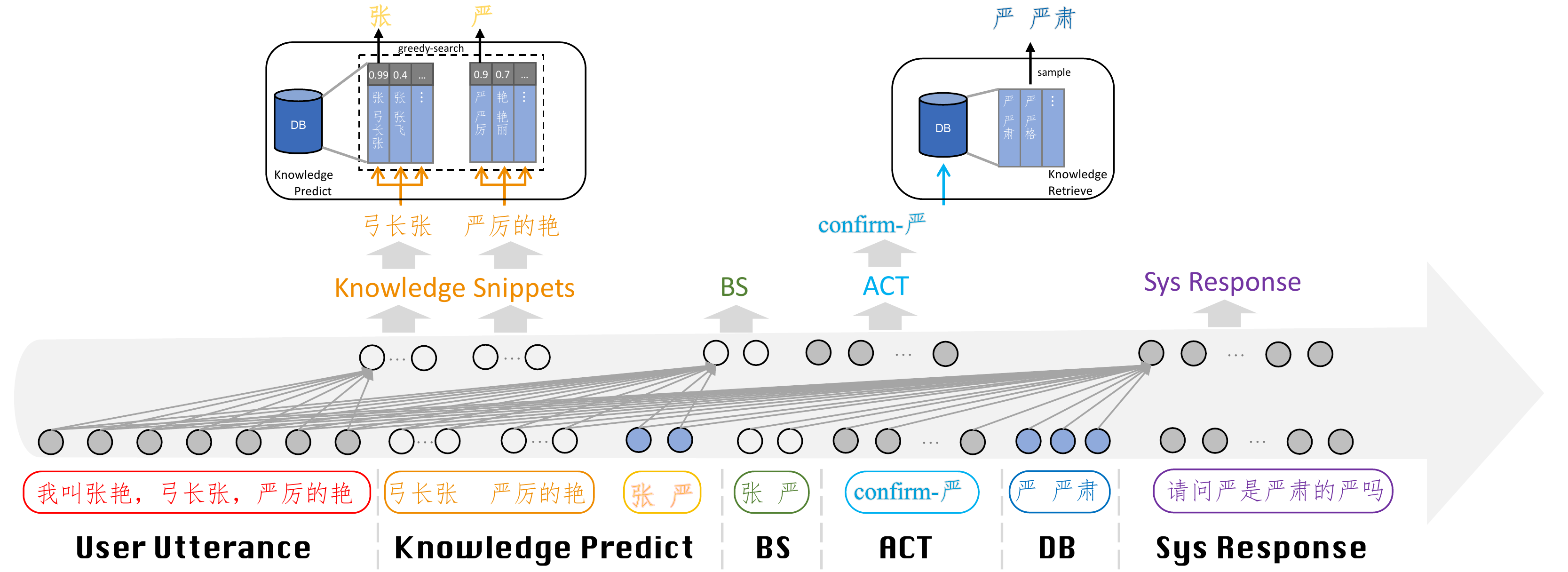}
    \caption{The structure of \our.}
    \label{fig: model}
\end{figure*}

Since the new task raises critical challenges, we firstly verify whether the current state-of-the-art (SOTA) models on normal task-oriented dialog task can meet the challenges, then we take a small step on improving one SOTA model by introducing a specific plug-in component to make it handle some of the challenges. 

Recently, many strong models have been proposed to tackle the MultiWOZ benchmark \cite{2020-simpleTOD, yang2021ubar, he2021galaxy}. In this paper, 
we chose three SOTA dialog models for our SSTOD evaluation as follows:

\paragraph{TRADE} ~\citep{wu-etal-2019-trade} utilizes the generative approach and copy-generator mechanism for slot filling tasks. We construct a complete dialog system using TRADE and a rule-based policy module as a baseline.

\paragraph{SimpleTOD} ~\citep{2020-simpleTOD}  uses a single, causal language model to aggregate dialog state tracking, policy deciding, and response generating a cascaded generator. Leveraging the large pre-trained model such as GPT2, SimpleTOD achieved competitive results on MultiWOZ.

\paragraph{UBAR} ~\citep{yang2021ubar} presents variants on \citet{ham-etal-2020-end, Peng-2020-soloist, zhang2020task} to parameterize the dialog system as an auto-regressive model. It models the task-oriented dialog system on a dialog session level, instead of using all user and system utterances as inputs. Conditioned on all previous belief state, system acts and response, UBAR is easier to make inference and planning in current turn and achieves the state-of-the-art performance on MultiWOZ.

\subsection{Plug-in Module}

As described above, one of the challenges in SSD is that the disambiguation of the slot values intensely relies on both the context and the extra knowledge. For example, users might inform a person name by making use of character knowledge to distinguish the target characters from alternatives. 

We therefore design a simple plug-in unit to execute Knowledge Prediction (KP) and Knowledge Retrieve (KR) on demand. Taking UBAR as a testbed, we proposed a UBAR with the plug-in unit (hereafter UBAR$^{+}$) whose framework is illustrated in Figure~\ref{fig: model}.

Given a user input utterance $U_t$, UBAR$^{+}$ first generates knowledge snippets $K_t=[k_{t}^1, ..., k_{t}^m]\subset U_t$, where $m$ is the number of extracted snippets. Each snippet corresponds to a target sub-slot value. For instance, if utterance $U_t$=``\begin{CJK*}{UTF8}{gbsn}我叫张艳，张是弓长张，艳是严厉的艳\end{CJK*}'', the extracted knowledge snippets $K_t=[k_{t}^1,k_{t}^2]$ = [``\begin{CJK*}{UTF8}{gbsn}弓长张 \end{CJK*}'', ``\begin{CJK*}{UTF8}{gbsn}严厉的艳 \end{CJK*}'']. 

Both extracted knowledge snippets and the knowledge items in extra knowledge base are embedded via TF-IDF \citep{jones1972statistical} vectors both in char-level and pinyin-level (which is the phonetic transcription of a Chinese character). 

Finally, the cosine similarities between the snippet $k_{t}^i\in K_t$ and each candidate knowledge item $kd_j$ from the knowledge base, are calculated as follows:

\begin{equation}
    e_c(k_{t}^i)=\text{TF-IDF}_{char}(k_{t}^i),
\end{equation}
\vspace{-0.5cm}
\begin{equation}
    e_p(k_{t}^i)=\text{TF-IDF}_{pinyin}(k_{t}^i),
\end{equation}
\vspace{-0.5cm}
\begin{align}
        score(k_{t}^i, kd_j)= \alpha & \cos{(e_c(k_{t}^i),e_c(kd_j))} \nonumber \\
    +(1-\alpha) & \cos{(e_p(k_{t}^i),e_p(kd_j))},
\end{align}
where $e_c(k_{t}^i)$ and $e_p(k_{t}^i)$ have the length of vocabulary size of characters and pinyin, respectively. 

For knowledge item $kd_k$ with the maximum similarity score, its corresponding character $w_k$ is used as the disambiguated character of $k_{t}^i$, yielding the predicted target sub-slot sequence $C_t=[w_1,\ldots,w_m]$.

Hereto we finish the disambiguation of one sub-slot value. By repeating the above procedures, all sub-slots are assigned their predicted target char, thereby the belief state (BS in Figure~\ref{fig: model}) is updated accordingly. To rationally navigate the following dialog, the agent then learns to plan its following acts of whether confirming a sub-slot or continuously requesting a sub-slot. 
We apply cross-entropy and language modeling objective \citep{bengio2003neural} to optimize the plug-in unit:
\begin{equation}
    L_{plug-in}=\sum_{i}\log{P(w_t|w_{<t})}.
\end{equation}
$L_{plug-in}$ is added to the loss applied in UBAR, making the final loss of the UBAR$^{+}$.

%% file: 4_Experiments.tex
\section{Experiments}

Using the \data dataset as a dialog state tracking benchmark, we conduct a comprehensive analysis of the challenges through an empirical approach and validate the effectiveness of the proposed UBAR$^+$ method.

\subsection{Experimental Setup}

\paragraph{Dataset.} We split the \data dataset into a training set, a validation set and a test set in the ratio of 7:1:2 on each of the four domains and conduct experiments on them.
\paragraph{Evaluation Metrics.} We evaluate model performances on \data with several popularly used metrics. \textbf{Joint acc} is the accuracy of all sub-slot values at each turn. The output is considered as an accurate one if and only if all the sub-slot values are exactly consistent with the ground truth values. \textbf{Slot acc} means whether each sub-slot is correctly collected at each turn. \textbf{Dialog succ} measures whether the collected slot value is consistent with the user’s goal at the end of the dialog. To have a comprehensive comparison, we also test our model by online interacting with FSA-based user simulators with two evaluation metrics: \textbf{Dialog succ} and \textbf{Avg turn}. \textbf{Dialog succ} is the main metric, which means the ratio of successful dialogs. A dialog is successful if and only if the slot is correctly collected by system within limited turns. \textbf{Avg turn} is used to measure the average turn number of successful dialogs.

\begin{table*}[t]
\small
\centering
\begin{tabular}{c|ccc|ccc|ccc|ccc} \hline
\multirow{2}{*}{Model}   & \multicolumn{3}{c}{\datap} & \multicolumn{3}{|c}{\datai}   & \multicolumn{3}{|c}{\datan}       & \multicolumn{3}{|c}{\datac} \\
                        & \tabincell{c}{Joint \\ acc} & \tabincell{c}{Slot \\ acc} & \tabincell{c}{Dialog \\ succ} & \tabincell{c}{Joint \\ acc} & \tabincell{c}{Slot \\ acc} & \tabincell{c}{Dialog \\ succ} & \tabincell{c}{Joint \\ acc} & \tabincell{c}{Slot \\ acc} & \tabincell{c}{Dialog \\ succ} & \tabincell{c}{Joint \\ acc} & \tabincell{c}{Slot \\ acc} & \tabincell{c}{Dialog \\ succ}  \\ \hline
TRADE*                  &  56.14  & 73.54  & 32.32    & 40.10   & 62.51    & 5.01     & 65.45      & 83.36    & 28.29         & 12.56     & 13.85    &  2.89    \\
SimpleTOD               &  \textbf{72.56}  & \textbf{85.80}  & \textbf{48.27}    & \textbf{70.17}   & \textbf{86.81}    & \textbf{43.50}    & \textbf{79.22}      & \textbf{91.24}    & \textbf{51.50}         & \textbf{48.55}     & 61.20    &  \textbf{36.58}        \\
UBAR                    &  71.62  & 85.23  & 46.00    & 69.70   & 86.60    & 40.70    & 63.58      & 82.58    & 34.40         & 47.70     &  \textbf{61.76}   &  35.20   \\ \hline

\end{tabular}
\caption{Comparisons of DST metrics and dialog succ on \data on the four domains.}
\label{offlinetest}
\end{table*}

\begin{table*}[t]
\small
\centering
\begin{tabular}{c|cc|cc|cc|cc} \hline
\multirow{2}{*}{Model}   & \multicolumn{2}{c}{\datap} & \multicolumn{2}{|c}{\datai}   & \multicolumn{2}{|c}{\datan}       & \multicolumn{2}{|c}{\datac} \\
                        & \tabincell{c}{Avg turn} & \tabincell{c}{Dialog succ} &  \tabincell{c}{Avg turn} & \tabincell{c}{Dialog succ} 
                         & \tabincell{c}{Avg turn} & \tabincell{c}{Dialog succ}  & \tabincell{c}{Avg turn} & \tabincell{c}{Dialog succ} \\ \hline
TRADE*                  &  9.77  &   30.45     &  16.68     &  26.39   &  6.75  &   5.71    &  6.50   &   20.26    \\
SimpleTOD               &  \textbf{8.18}  & \textbf{63.20}      & \textbf{10.94}       &  \textbf{46.70} & 4.79 & \textbf{15.80}  & \textbf{6.29} & \textbf{32.70} \\
UBAR                    &  11.39  & 57.7 & 10.97 & 41.50 & \textbf{4.41} & 11.50 & 6.63  & 25.10  \\ \hline

\end{tabular}
\caption{Results of different models on interaction with a FSA-based user simulator on four domains.}
\label{onlinetest}
\end{table*}

\paragraph{Implementation Details.} We initialize our proposed UBAR$^+$ model with ClueCorpus-small~\citep{xu2020cluecorpus2020} and fine-tune it on \data. The max length of an input sequence is set to $1024$ and the excess parts are truncated. The $\alpha$ in the plug-in unit is set to $0.09$. AdamW~\citep{loshchilov2018AdamW} optimizer is applied and the learning rate is initialized as $0.0001$. 

\subsection{Results and Analysis}

We implement three different evaluations on model performances: The first one is offline test where models are evaluated using \data test data, the second one is online test where models interact with FSA-based user simulator, and the third one is human evaluation where models interact with humans. 

The offline evaluation results of the three baseline models across all domains on \data are summarised in Table~\ref{offlinetest}. As we can see, all three models perform poorly, and nearly all the dialog success rates are lower than $50$\%. Remind that the success rate of UBAR on MultiWOZ is higher than $70$\%. Among them, GPT2 based models (SimpleTOD and UBAR) achieve relatively good performance on \data owing to the efficacy of large pre-trained language models. Although SimpleTOD achieves the best results on all four domains. Nevertheless, SimpleTOD only reaches nearly $40$\% dialog success on \datap and \datai, $51.50$\% on \datan, and $36.58$\% on \datac.  
Table~\ref{onlinetest} illustrates the results of online evaluations. The similar observations are concluded as those in offline evaluations. Even the most efficient SimpleTOD model achieves poor success rates.

\begin{figure}
    \centering
    \includegraphics[width=\linewidth]{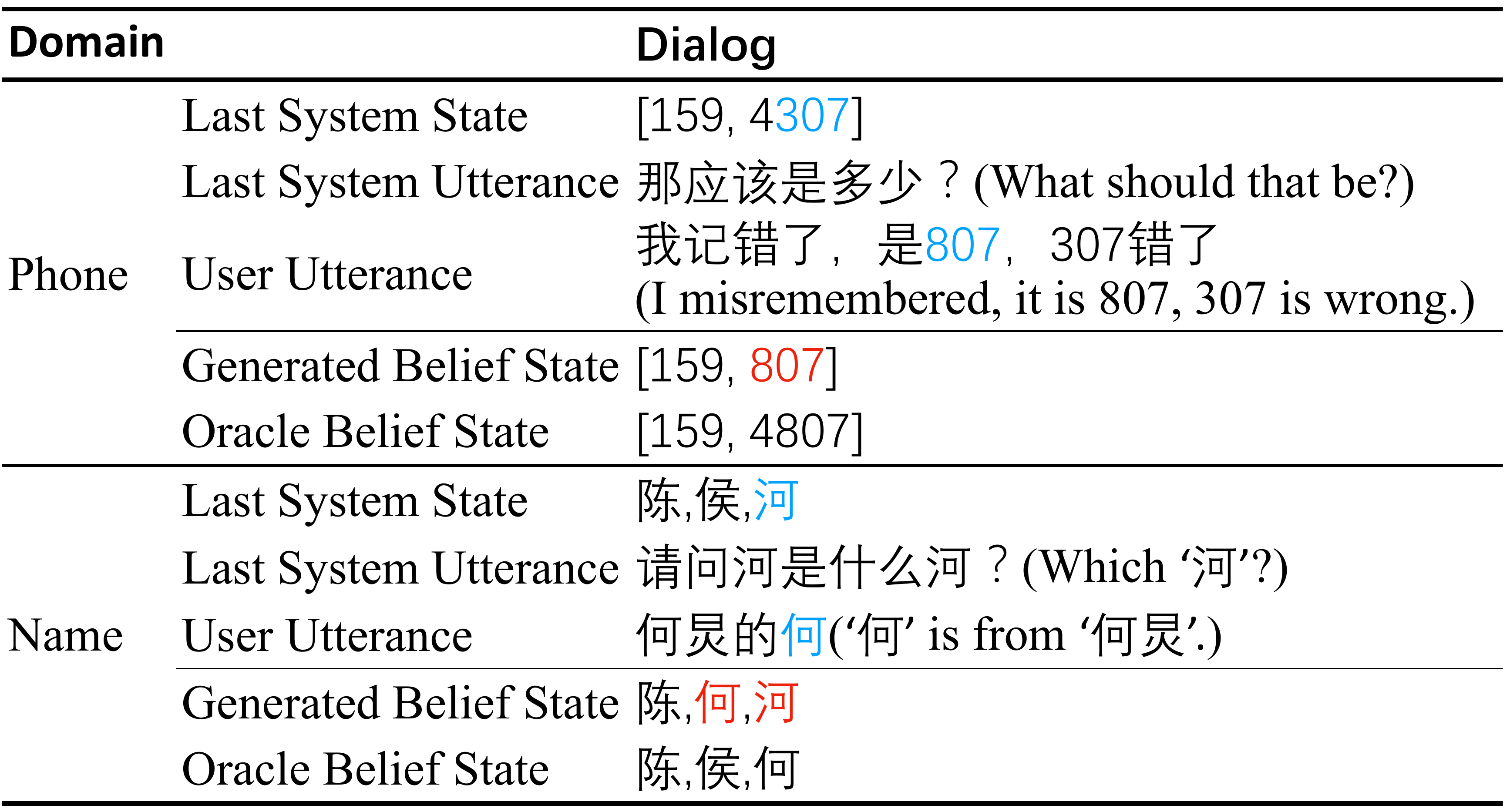}
    \caption{Typical bad cases of UBAR. In the phone domain, the system ought to update part of the second sub-slot ``307'' to ``807'' but it updates the whole sub-slot by mistake. In the name domain, system indexes a wrong sub-slot ``\begin{CJK*}{UTF8}{gbsn}侯\end{CJK*}'' and changes it to ``\begin{CJK*}{UTF8}{gbsn}何\end{CJK*}''.}
    \label{fig:badcase}
\end{figure}

From the detailed analysis of the results, we observe that one of the major factors affecting the performance is the difficulty of sub-slot locating, especially when updating a fragment of the sub-slot. In the phone number domain and ID number domain, the system should compare the updated fragment with the collected value to determine which fragment is similar to that one. As shown in Figure~\ref{fig:badcase}, the system is required to change ``307'' to ``807'', but it wrongly updates ``4307'' to ``807''. 
For the name slot, the system changes ``\begin{CJK*}{UTF8}{gbsn}侯\end{CJK*}'' to ``\begin{CJK*}{UTF8}{gbsn}何\end{CJK*}'' by mistake. When taking the ASR noise into account, the scenarios would become much more complicated.


\subsection{Performance of plug-in unit}

Table~\ref{our approach} shows the performance of our knowledge plug-in unit on \datan. \our performs the best, with $23$\% improvement over UBAR and $6$\% improvement over SimpleTOD in terms of dialog succ. We claim that the knowledge plug-in unit enables the model to obtain relevant knowledge by querying the knowledge base, which is beneficial to complete slot value acquisition and response generation. 

Further investigation is conducted through interaction between the model and the user simulator. Table~\ref{our approach} shows \our harvests a great improvement in name collecting, yielding an accuracy rate of $45.8$\%, which further proves the efficiency of knowledge-rich disambiguation. The same trend is also observed for the other three domains.

\begin{table}[t]
    \small
    \centering
    \begin{tabular}{c|ccc|cc} \hline
    \multirow{2}{*}{Model} & \multicolumn{3}{c}{Offline Test} & \multicolumn{2}{|c}{Online Test} \\
                           & \tabincell{c}{Joint \\ acc} & \tabincell{c}{Slot \\ acc} & \tabincell{c}{Dialog \\ succ} & \tabincell{c}{Avg \\ turn} & \tabincell{c}{Dialog \\ succ} \\ \hline
SimpleTOD    & 79.22      & 91.24    & 51.50  & 4.79 & 15.80 \\
UBAR         & 63.58      & 82.58    & 34.40  & \textbf{4.41} & 11.50 \\ \hline
\our    & \textbf{84.96}      &  \textbf{93.12}   & \textbf{57.73}  & 4.60 & \textbf{45.80} \\ \hline
    \end{tabular}
    \caption{Comparisons between \our and the SOTA models in both offline and online tests on the Chinese name domain.}
    \label{our approach}
\end{table}

\subsection{Human Evaluation}

\begin{table}[ht]
\centering
\small
\begin{tabular}{c|ccc} \hline
Model & Dialog succ & App & Diversity \\ \hline
UBAR  & 28.00 & 2.82 & 3.10 \\
\our  & \textbf{50.00} & \textbf{2.89} & \textbf{3.96} \\ 
 \hline
\end{tabular}
\caption{Performance on human evaluation on Chinese name domain. App indicates the average appropriateness scores.}
\label{human result}
\end{table}

For human evaluation, $10$ postgraduates are recruited to evaluate \our and UBAR on Chinese name domain. During the interaction, the students randomly change the characters to those with similar pronunciations in the sentences. The same name and knowledge with errors are used on both models. At the end of the conversation, the evaluators are asked to check whether the dialog is successful. The postgraduates also score each system response to evaluate the appropriateness of the system response~\cite{zhang2020task}. The points range from 1 to 3, which respectively represent \emph{invalid}, \emph{ok}, and \emph{good}. Another score on a Likert scale of 1-5 evaluates the diversity of the whole dialog. The results are shown in Table~\ref{human result} and prove that \our yields a much higher dialog success rate.

%% file: 6_Related_Work.tex
\section{Related Work}

We can group the datasets for task-oriented dialog systems by whether the two parts involved in the dialogs are humans or machines: human-to-human (H2H), machine-to-machine (M2M) and human-to-machine (H2M) collecting methods. H2H corpora are derived by asking a human user to talk with a human agent. To mimic the conversations between human and machine, H2H datasets ubiquitously apply the Wizard-of-Oz approach~\cite{hemphill-etal-1990-atis,el-asri-etal-2017-frames,budzianowski-etal-2018-multiwoz,zhu-etal-2020-crosswoz}, which a human agent pretends as machine to talk to a human user and the human user believes the other side is a machine. However, it costs tremendous effort to construct such a H2H dataset. M2M datasets which are generated by simulated systems and simulated users take much less work to construct than H2H datasets with the same scale. However, the naturalness and diversity of M2M datasets are questioned~\cite{peng-etal-2017-composite,shah2018building,rastogi2020towards, dai-etal-2020-learning}. 
H2M~\cite{raux2005let,williams-etal-2013-dialog,henderson-etal-2014-dstc2,henderson-etal-2014-word,dstc4} hires crowd workers to chat with a machine system and the conversations are more diverse and natural than M2M. We integrate the M2M and H2M approaches by boosting the generated M2M datasets through crowdsource rewriting to obtain more diverse and natural dialogs with less effort. 

The datasets might be also grouped by the single-domain and the multi-domain. The early datasets are mostly single-domain. For example, ATIS~\cite{hemphill-etal-1990-atis}, by M2M strategy, is a system to help people make air travel plans; a H2M corpus, Let’s Go Public~\cite{raux2005let}, contains consultation dialogs of bus schedule information; two datasets for buying a movie ticket and reserving a restaurant table are collected by M2M~\cite{shah2018building}. Single-domain systems generally fill slots within a single turn and thereby slot values are relatively independent. Recently, multi-domain datasets grab more attention. MultiWOZ~\cite{budzianowski-etal-2018-multiwoz}, one of the most popular datasets, consists of Wizard-of-Oz large-scale multi-domain conversations. A M2M dataset, SGD~\cite{rastogi2020towards}, generates multi-domain dialogs, guided by the predefined schema. CrossWOZ~\cite{zhu-etal-2020-crosswoz} states how slots in one domain relate to the following domains by reference. Nevertheless, none of the above datasets, with single domain or multiple domains, look into sub-slot cases as \data does. In \task, we have to not only locate the related previous sub-slots through complicated expressions, but also tile the pieces of value into a correct sequence without duplication, missing, and errors under the assistance of external knowledge. 

%% file: 7_Conclusion.tex
\section{Conclusions and Future Work}

In this paper, we propose a sub-slot based task \task which has not brought to the public. To help the exploration of the task, we build a textual dialog dataset \data which covers four popular domains and contains natural noise brought by ASR module. 
 \data stems from the real human-to-human dialogs and can be utilized as a benchmark for slot filling, dialog state tracking and dialog system that matches the real-world scenarios.

\section*{Ethical Considerations}
The collection of our \data dataset is consistent with the terms of use of any sources and the original authors’ intellectual property and privacy rights.
The \data dataset is collected with ALIDUTY\footnote{https://www.aliduty.com/} platform, and each HIT requires up to 10 minutes to complete. The requested inputs are general language variations, speech voices, and no privacy-related information is collected during data collection. Each HIT
was paid 0.1-0.2 USD for a single turn dialog data, which is higher than the minimum wage requirements in our area. The platform also hires professional reviewers to review all the collected data to ensure no ethical concerns e.g., toxic language and hate speech.

\section*{Acknowledgements}
The authors would like to thank anonymous reviewers for their suggestions and comments, as well as professors for great inspiration. The authors are also grateful for all teammates for their productive and constructive advice.

%% file: 8_Appendix.tex
\section{Knowledge}
\label{sec:appendix1}

\begin{figure}[ht]
    \centering
    \includegraphics[width=\linewidth]{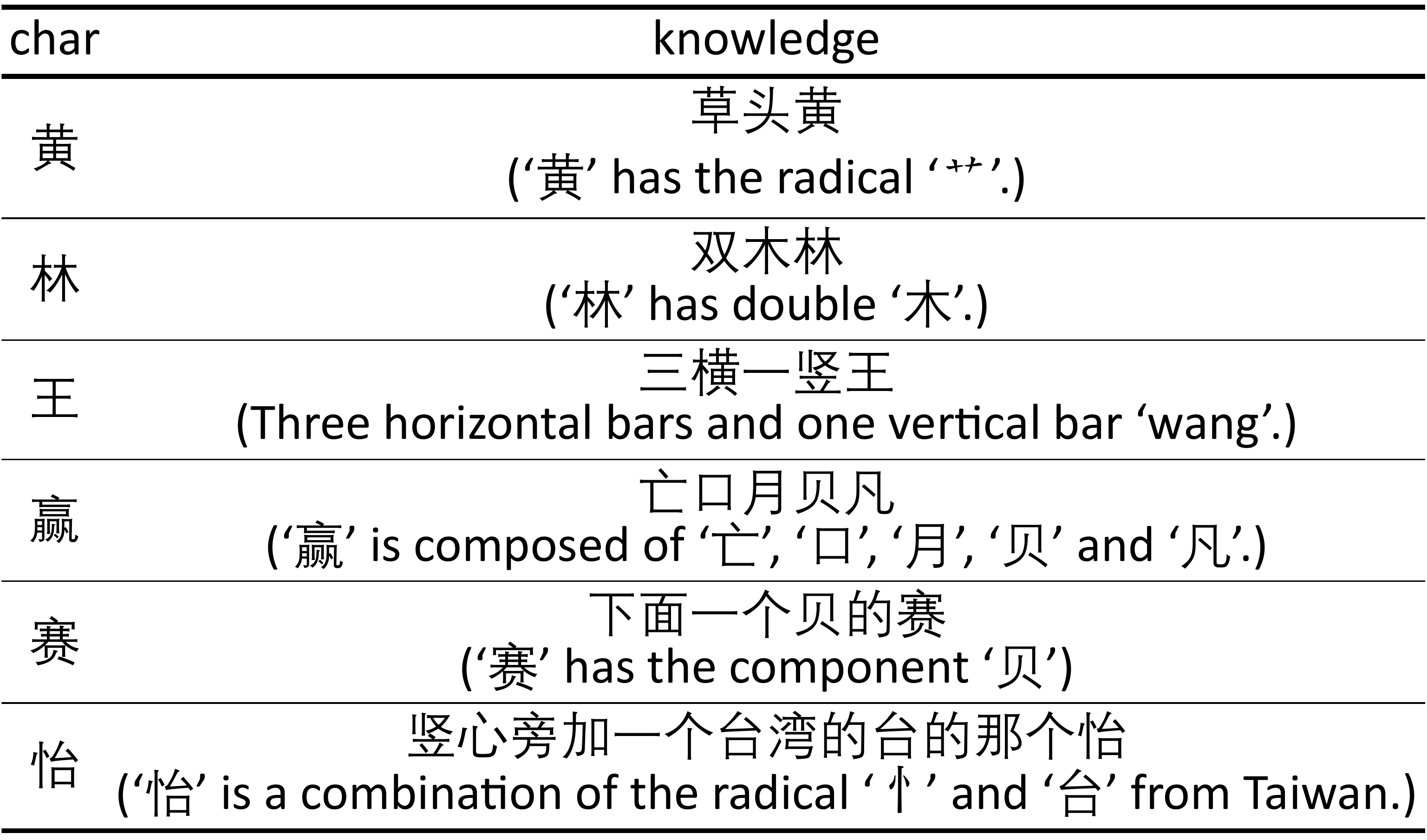}
    \caption{Some different types of knowledge on Chinese characters.}
    \label{fig:different types of knowledge}
\end{figure}

\begin{figure}[ht]
    \centering

    \includegraphics[width=\linewidth]{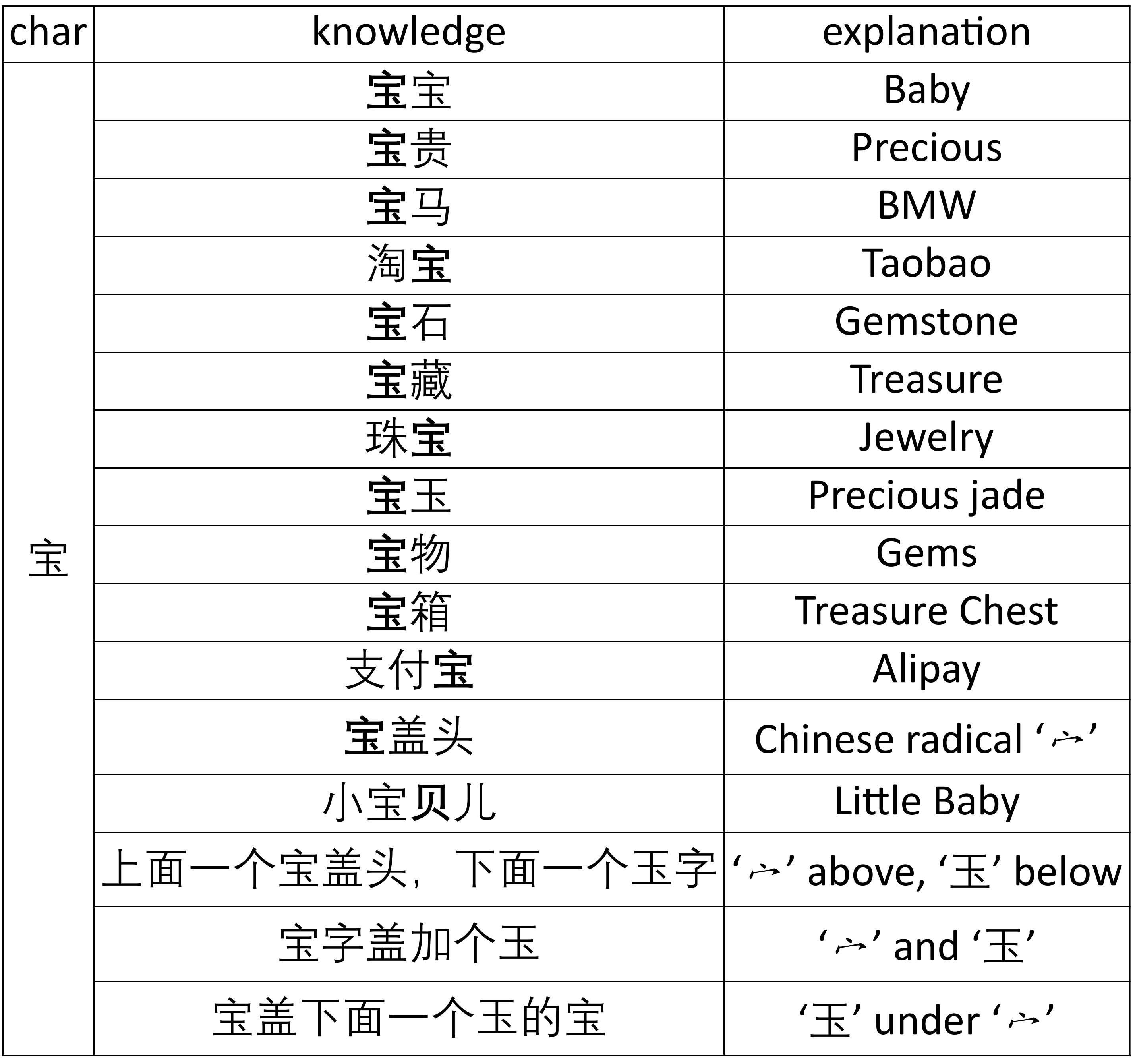}
    \caption{Some pieces of knowledge about \begin{CJK*}{UTF8}{gbsn}宝\end{CJK*}'.}
    \label{fig:knowledges of 'bao'}
\end{figure}

Figure~\ref{fig:different types of knowledge} shows some different types of knowledge. The word ``\begin{CJK*}{UTF8}{gbsn}黄\end{CJK*}'' is described with its radical. And it is necessary to use whole components to explain ``\begin{CJK*}{UTF8}{gbsn}林\end{CJK*}'', ``\begin{CJK*}{UTF8}{gbsn}王\end{CJK*}'' and  ``\begin{CJK*}{UTF8}{gbsn}赢\end{CJK*}'' . In the fourth row, only a part of the word ``\begin{CJK*}{UTF8}{gbsn}赛\end{CJK*}'', ``\begin{CJK*}{UTF8}{gbsn}贝\end{CJK*}'', is enough to disambiguate homonyms. In the last example, ``\begin{CJK*}{UTF8}{gbsn}台\end{CJK*}'' also needs explanation besides ``\begin{CJK*}{UTF8}{gbsn}怡\end{CJK*}''. Overall, the knowledge description is challenging for systems to get the correct char. 

Some pieces of knowledge about ``\begin{CJK*}{UTF8}{gbsn}宝\end{CJK*}'' are shown in Figure~\ref{fig:knowledges of 'bao'}. It contains phrase-based knowledge, structure-based knowledge and hybrid knowledge. The way to explain one character is various and the number of one character's knowledge is large.

\newpage

\section{NLG Templates}
\label{sec:appendix2}
\begin{figure}[ht]
    \centering
    \includegraphics[width=\linewidth]{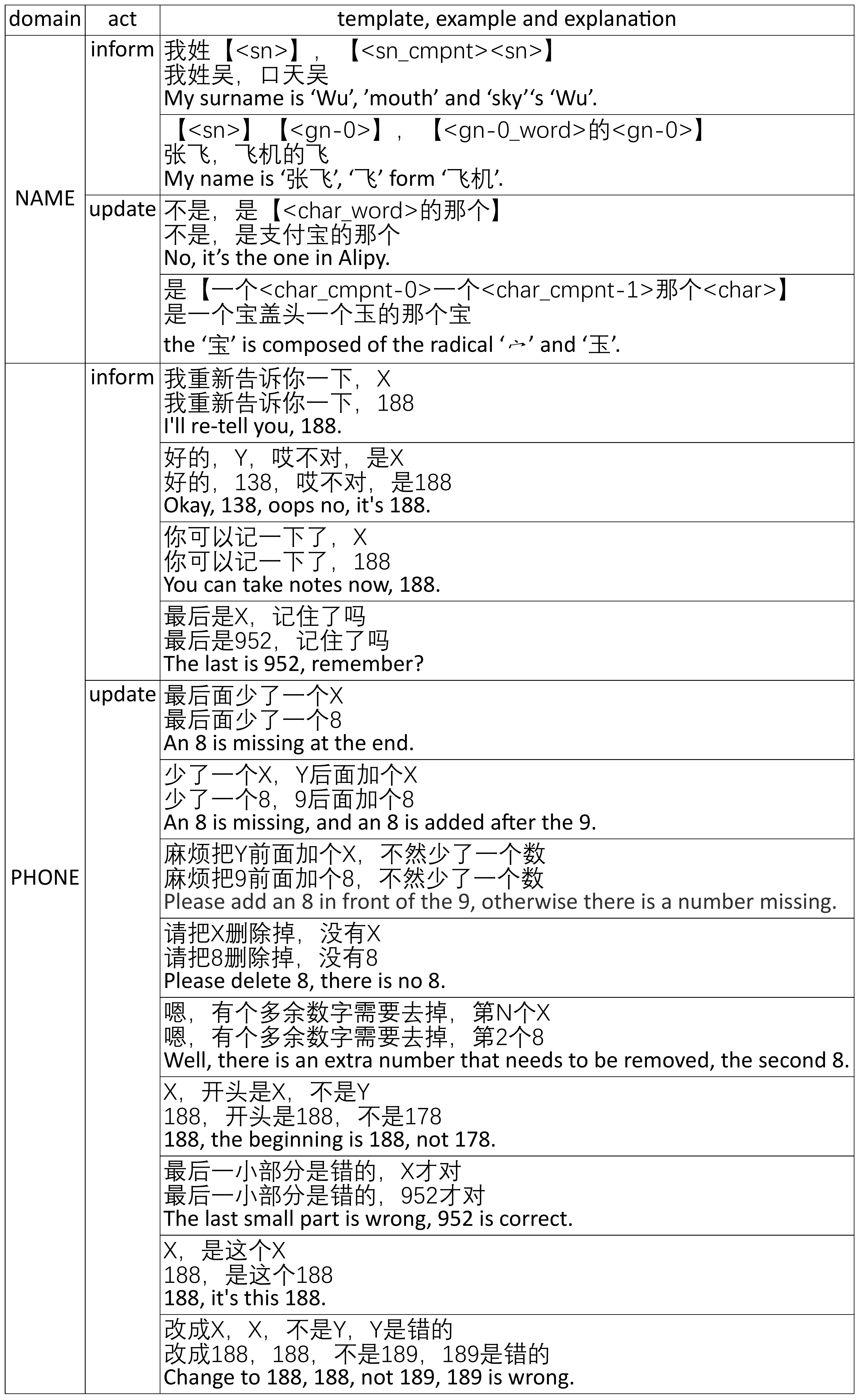}
    \caption{Some examples of NLG templates. Each cell in the third line is template, a sentence example generated by the template, the explanation of the generated sentence.}
    \label{fig: NLG templates}
\end{figure}

Some NLG templates are presented in Figure~\ref{fig: NLG templates}. In the domain of name, Chinese name consists of surname ``<sn>'' and given name ``<gn>''. Each word in name has two kinds of knowledge, components ``<\_cmpnt>'' and words ``<\_word>'', to distinguish different characters.
In the phone domain, when generating one sentence using a template, `X' is replaced by a sub-slot value to be informed and `Y' is replaced by the noisy sub-slot value or which to be updated. 

\newpage

\section{FSA}
\label{sec:appendix3}

\begin{figure}[ht]
    \centering
    \includegraphics[width=\linewidth]{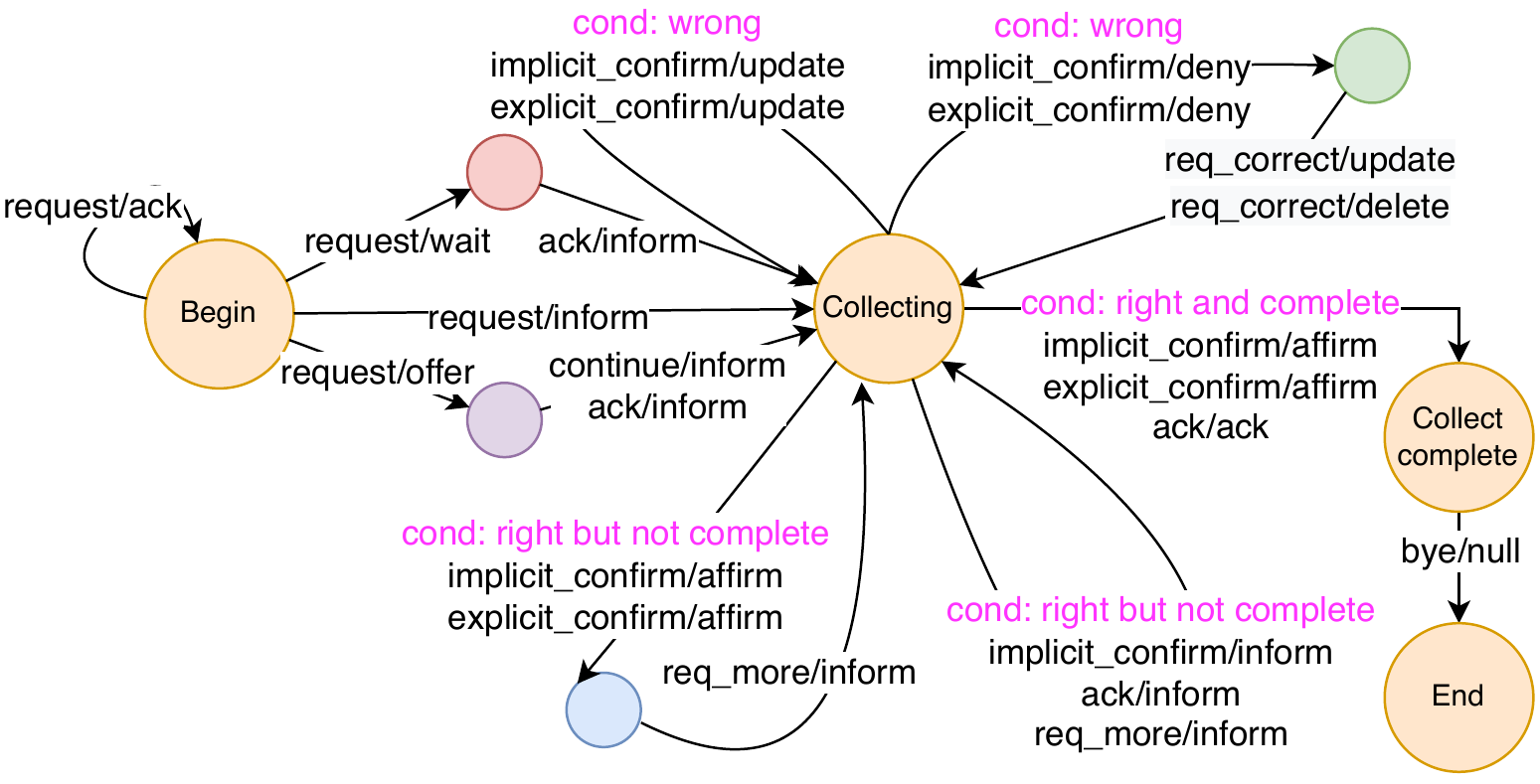}
    \caption{Part of automata that generates the dialog of \datap dataset and \datai dataset.}
    \label{fig:fsa}
\end{figure}

Taking the construction process of \datap dataset as an example, the partial automata constructed by analyzing the H2H dialog is shown in Figure~\ref{fig:fsa}. When a system act is chosen, the user transfers state according to the user state and the system slot value and outputs the user act with probability. At the same time, in order to imitate various user expressions in the real world, one user act can correspond to various expressions as shown in Table~\ref{tbl:diversity-example}.

\newpage

\section{Dialog example}
\label{sec:appendix4}
\begin{figure}[ht]
    \centering
    \includegraphics[width=\linewidth]{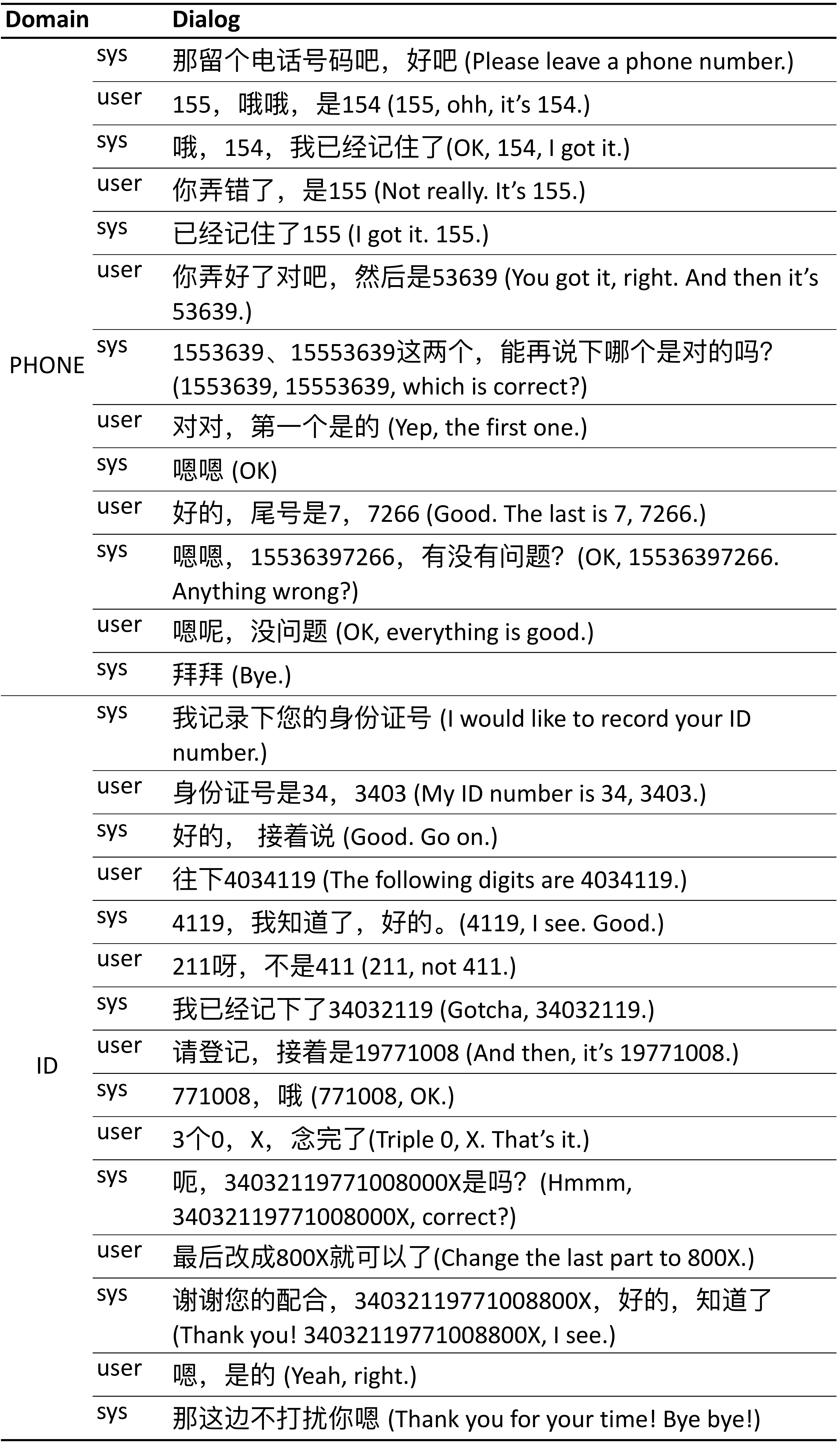}
    \caption{Examples of dialogs of phone number collection and identity number collection.}
\end{figure}

\newpage

\begin{figure}[ht]
    \centering
    \includegraphics[width=\linewidth]{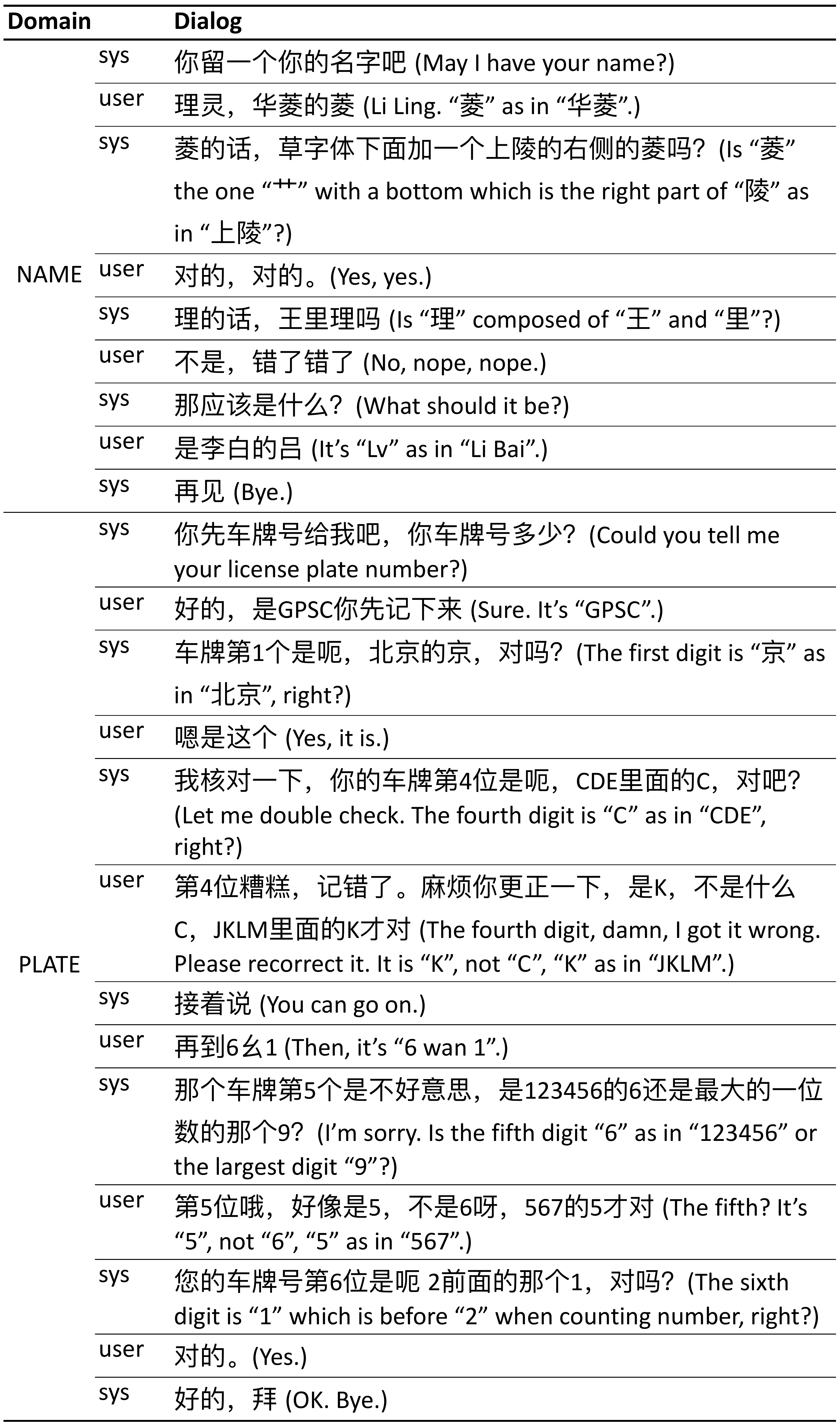}
    \caption{Examples of dialogs of name collection and license plate number collection.}
\end{figure}